\documentclass[sigconf]{acmart}
\pdfoutput=1
\usepackage{booktabs} % For formal tables
\usepackage[utf8]{inputenc}
\usepackage{color}
\usepackage{mathtools}
\usepackage{balance}
\newcommand*{\suchthat}[1]{\left|\vphantom{#1}\right.}
\DeclareMathOperator*{\argmin}{arg\,min}  % in your preamble 
\setcopyright{rightsretained}
\acmDOI{10.475/123_4}

\acmISBN{123-4567-24-567/08/06}

\acmConference[ICDSC'17]{International Conference on Distributed Smart Cameras}{September 2017}{Stanford, California USA}
\acmYear{2017}
\copyrightyear{2016}

\acmPrice{15.00}

\begin{document}
\title{A location-aware embedding technique for accurate landmark recognition}
%\titlenote{Produces the permission block, and copyright information}
%\subtitle{Extended Abstract}
%\subtitlenote{The full version of the author's guide is available as
%  \texttt{acmart.pdf} document}

\author{Federico Magliani}
\orcid{0000-0001-5526-0449}
\affiliation{%
  \institution{IMP lab - D.I.A.\\ Università di Parma}
  \city{Parma} 
  \state{Italy} 
}
\email{federico.magliani@studenti.unipr.it}

\author{Navid Mahmoudian Bidgoli}
%\authornote{This author is the one who did all the really hard work.}
\affiliation{%
  \institution{INRIA Rennes Bretagne-Atlantique}
  %\streetaddress{Campus de Beaulieu, 263}
  \city{Rennes} 
  \country{France}
  %\postcode{35042}
    }
\email{navid.mahmoudian-bidgoli@inria.fr}

\author{Andrea Prati}
\affiliation{%
  \institution{IMP lab - D.I.A.\\ Università di Parma}
  %\streetaddress{Parco Area delle Scienze, 181/A}
  \city{Parma} 
  \state{Italy} 
  %\postcode{43124}
}
\email{andrea.prati@unipr.it}

% The default list of authors is too long for headers}
\renewcommand{\shortauthors}{F. Magliani et al.}

\begin{abstract}
%Currently a global descriptor, composed by local aggreagted features, called VLAD reaches good results on the object recognition problem.
%Focusing on this approach, we propose a new technique on the VLAD pipeline for incrementing the accuracy results on popular benchmarks for the intra-dataset problem. Considering that, usually, at the border of an image there are many noisy features like cars, street, people; optVLAD try to reduce this problem calculating a mean of two global descriptors: the VLAD descriptor of the original image and the VLAD descriptor of the cropped image. This operation make benefits if it is applicated only on the query images. Furthermore we dedrimonstrate, through the use of graphics, that we can obtain the same results with VLAD using a reduced number of features during the initial clustering phase.

The current state of the research in landmark recognition highlights the good accuracy which can be achieved by embedding techniques, such as Fisher vector and VLAD.
All these techniques do not exploit spatial information, i.e. consider all the features and the corresponding descriptors without embedding their location in the image.
This paper presents a new variant of the well-known VLAD (Vector of Locally Aggregated Descriptors) embedding technique which accounts, at a certain degree, for the location of features. The driving motivation comes from the observation that, usually, the most interesting part of an image (e.g., the landmark to be recognized) is almost at the center of the image, while the features at the borders are irrelevant features which do no depend on the landmark. The proposed variant, called locVLAD (location-aware VLAD), computes the mean of the two global descriptors: the VLAD executed on the entire original image, and the one computed on a cropped image which removes a certain percentage of the image borders.
This simple variant shows an accuracy greater than the existing state-of-the-art approach.
%, but also allows to reduce the number of features used during the clustering and still obtain accuracy comparable with existing methods.
Experiments are conducted on two public datasets (ZuBuD and Holidays) which are used both for training and testing. Morever a more balanced version of ZuBuD is proposed.
\end{abstract}

%
% The code below should be generated by the tool at
% http://dl.acm.org/ccs.cfm
% Please copy and paste the code instead of the example below. 
%
%\begin{CCSXML}
%<ccs2012>
%<concept>
%<concept_id>10010147.10010178.10010224.10010245.10010251</concept_id>
%<concept_desc>Computing methodologies~Object recognition</concept_desc>
%<concept_significance>500</concept_significance>
%</concept>
%</ccs2012>
%\end{CCSXML}
%
%\ccsdesc[500]{Computing methodologies~Object recognition}

% We no longer use \terms command
%\terms{Theory}


\maketitle

\section{Introduction}
Mobile landmark recognition is an interesting research field of computer vision. 
%This problem isn't currently been solved in all possible tasks, but now many its objectives are reachable with good results because of the increasing in developing small hardware components at low price with most power than in the recent past.
Basically, it consists of a client-server application: the client (e.g. a mobile device) sends a picture of a place to the server, that tries to recognize the place (or landmark) in a fast way and sends back the final result to the client. Possible applications of mobile landmark recognition range from augmented reality with information about the landmark, to image-based geo-localization of the device, to advanced electronic tourist guides.

Generally speaking, landmark recognition is a challenging task and, therefore, it is still a very active field of research. Among the possible challenges, those related to the differences of two images of the same place are the most relevant for computer vision algorithms. Changes in the image resolution, illumination conditions, viewpoint and the presence of distractors such as trees or traffic signs (just to mention some) make the task of matching features between a query image and the database rather difficult.
In order to mitigate these problems, the existing approaches rely on feature description with a certain degree of invariance to scale, orientation and illumination changes.

%To evaluate the approach proposed in the image recognition problem there are public image dataset, that are ordinarily divided in two folders: database and query images.
%The database images are composed by some pictures for every class; in which every class represents a different place.
%Instead the query set is composed, usually, by one image for every class, different from the database images of the relative class. The query images are resized, rotated, taken as photos in different view or position compared to the database images of the same classes.
%The complexity of the dataset is proportional to the summatory of differences between the query and database images of the same class.

From the experimental perspective, in the field of landmark recognition (as well as in other similar fields) it is common (and often mandatory) to use public datasets, with the clear advantage to have a fair and immediate comparison with competitive approaches. Two experimental setups are possible: the first in which the training of vocabulary words and the testing (or query) images come from the same dataset (often called \textit{intra-dataset} setup); the second in which training is performed on one dataset, whereas the query images belong to another dataset (\textit{inter-dataset} setup). The second setup aims at demonstrating the generalization property of the proposed approach.

This paper introduces the following novel contributions on the landmark recognition problem:
\begin{itemize}
\item a location-aware version of VLAD, called locVLAD, that allows to outperform the state of the art in the intra-dataset problem;
\item the proposed locVLAD technique is descriptive and discriminant enough to achieve an accuracy comparable with the state of the art also when the number of features used during the vocabulary creation phase is significantly reduced, therefore speeding up the computation;
\item a new balanced version of the public dataset ZuBuD is proposed and made available to the scientific community; the new version represents equally all the classes in the dataset, by resulting in higher accuracy in the recognition process.
\end{itemize}

This paper is organized as follows. Section 2 introduces the techniques used in the state of the art. Section 3 briefly reviews VLAD and describes our new implementation of VLAD. Section 4 evaluates our methods on public benchmarks: ZuBuD and Holidays. Finally, concluding remarks are reported.

\section{Related work}

The Bag of Words (BoW) model is the first technique implemented for solving the problem of object recognition \cite{sivic2003video}. It is based on the creation of a vocabulary of visual words using a clustering algorithm applied on the training set. Then, each image during the testing phase is described in terms of occurence of these words. Though quite simple, BoW has achieved good results in image retrieval, at the cost, however, of a large consumption of memory.

Given the limitations of this approach, researchers have started to use vocabulary tree of descriptors \cite{chen2009tree, nister2006scalable}. It is an optimization of the BoW model for the representation of the features. 
Although the performance is improved with respect to BoW approach, this method required much more memory on the device therefore it is not applicable for mobile devices.
%This method required much memory on the device, needed a certain amount of computational time to insert new features in the tree and did not obtain excellent results. 

%Given the limitations of this approach, researchers have started to use Bag of Words (BoW) for image retrieval. BoW \cite{sivic2003video} is based on the creation of a vocabulary of visual words using a clustering algorithm applied on the training set. Then, each image during the testing phase is described in terms of occurence of these words. Though quite simple, BoW has achieved good results in image retrieval, at the cost, however, of a large consumption of memory.

To overcome the weakness of the BoW approach, several embedding techniques have been proposed in the literature. The first proposal on this direction has been done by Perronnin and Dance in \cite{perronnin2007fisher}: here, Fisher Kernels are used to encode the vocabularies of visual words represented by means of a GMM (Gaussian Mixture Model).
Another well-known embedding technique is VLAD (Vector of Locally Aggregated Descriptors) \cite{jegou2010aggregating} which encodes the residual of a feature instead of the values of the features detected in the images. Given its simplicity of implementation and the good results achievable, VLAD is very diffused and several variants have been proposed in the literature. For instance, the CVLAD (Covariant VLAD) \citep{zhao2013oriented} creates different VLAD vectors for every orientation of the keypoints. CVLAD resulted in good recognition performance but at the cost of a large number of feature needed (due to the separation in different vectors which require enough data to be constructed).
In fact, the paper \citep{zhao2013oriented} employed a dense SIFT detector for obtaining the features from the images.

Zhang \textit{et al.} \cite{zhang2015probabilistic} implemented a method based on sparse coding and by using max pooling in alternative to sum pooling (used in the traditional VLAD implementation). 
%Furthermore the use of the color, based on the histogram calculated on the image, improved the results. Despite the good results achieved by this approach, reliable color information is not always available in real conditions.

An alternative embedding technique is represented by Hamming embedding \citep{jegou2008hamming}. Jégou \textit{et al.} binarized the values of descriptors detected in the images and calculated the similarites through the Hamming distance. Given its simplicity, this method can work well also on large datasets. Unfortunately, this approach is prone to the problem of burstiness, i.e. the presence of repetitive features in the image (which can be quite common, for example, in bricks of the building walls) can affect the value of the binarized descriptor significantly.

In the last years, with the new developments of powerful GPUs, the neural networks have allowed to resolve complex problems with good results. Sharif \textit{et al.} \cite{sharif2014cnn} used a CNN with an SVM classifier to solve the problem of image recognition. CNNs have been applied to solve many computer vision and machine learning problems, always reaching excellent results. This is also the case of landmark recognition. However, CNNs have two main drawbacks. The first is that they need a lot of data to be trained effectively, and this can be a challenging task in some cases. Secondly, the computational resources necessary for CNNs still make them hard to be implemented on mobile devices. Gong \textit{et al.} \cite{gong2014multi} implemented CNN, that makes use of VLAD embedding in several phases of the system. The results of each level are then pooled in the final descriptor.

With regards to the specific application, several previous works have been reported. The paper of Fritz \textit{et al.} \cite{fritz2006mobile} implemented a new version of the SIFT detector, called i-SIFT, that achieves a significant speedup by applying a filtering to remove the less promising candidate descriptors. The landmark recognition application developed by Chen \textit{et al.} \cite{chen2011city} used a vocabulary tree and RANSAC method for geometric verification of the top candidates. However, it requires some time to obtain the final results. Finally, Schroth \textit{et al.} \cite{schroth2011rapid} proposed an approach based on multiple hypotesis vocabulary tree.

\section{The locVLAD approach}

Before starting to describe the locVLAD variant, let us introduce the basic concepts of VLAD.
VLAD is based on computing a compact descriptor based on the residuals of feature descriptors. The original VLAD proposal \cite{jegou2010aggregating} uses an Hessian-Affine feature detector and SIFT as descriptor. Arandjelovic and Zisserman \citep{arandjelovic2012three} introduced a variant which substitutes SIFT descriptor with the so-called RootSIFT. This descriptor applies square root to the positive components of the descriptor and then a L\textsubscript{2} normalization is performed.

\textbf{Creation.} The first step for computing VLAD is the creation of the vocabulary. Let $k$ be the size of the $C = \{\mathbf{\mu_{1}, \dots, \mu_{k}}\}$ vocabulary (i.e., the number of visual words retained), then K-means clustering algorithm can be used on all the features in the training set to compute the cluster centers $\mathbf{\mu_i}$.

Once the vocabulary has been created, in the testing phase each of the $m$ descriptors extracted from the query image can be assigned to the closest cluster center. Being $X = \{\mathbf{x_1, \dots, x_m}\}$ the set of descriptors, the assignment function $q$ can be written as:

$$q : \mathbb{R}^d \to C$$

$$ q(\mathbf{x}) = \mathbf{\mu_i} \suchthat* i = \argmin_{i = 1, \dots, k} ||\mathbf{x-\mu_i}||$$

\noindent where $|| \cdot ||$ is a proper $d$-dimensional distance measure and $d$ is the size of the descriptors ($d$ = 128 in the case of SIFT). Each descriptor $\mathbf{x_j}$ is thus composed of $d$ value $x_j$ with $s = 1,\dots, d$.

The VLAD vector $\mathbf{v}$ is obtained by accumulating the residuals computed by the difference between the feature descriptor and the relative cluster center. Different strategies have been proposed in the literature. The two most common are the original \textit{sum aggregation} and the so-called \textit{mean aggregation} proposed in \cite{chen2013residual}. However, as shown by Spyromitros \textit{et al.} in \cite{spyromitros2014comprehensive} and confirmed by our tests, the sum aggregation yield the best results: therefore, defining $\mathbf{v} = \{\mathbf{v}^1, \dots, \mathbf{v}^k\}$ we can obtain the values of VLAD vector as follows:

$$ \mathbf{v}^i = \sum\limits_{\forall \mathbf{x} \in X : q(\mathbf{x}) = \mathbf{\mu}_i}{\mathbf{x} - \mathbf{\mu}_i} $$

Finally, the resulting $k \times  d$ vectors are concatenated to form the unnormalized VLAD vector $\mathbf{v}$.

\textbf{Normalization. } Several possible normalization strategies have been proposed in the past, such as power-law normalization \cite{perronnin2010improving} that updates the VLAD components using a power law, or intra normalization \citep{arandjelovic2013all}, that normalizes the sum of residuals of each block with L\textsubscript{2} normalization, or signed square rooting \citep{jegou2012aggregating}, where the VLAD components are updated with the absolute value of the square root of the element. 

However, it has been demonstrated in \cite{delhumeau2013revisiting} that the residual normalization results to be the best performing:

$$ \mathbf{\dot{v}}^i = \sum\limits_{\forall \mathbf{x} \in X : q(\mathbf{x}) = \mathbf{\mu}_i}{\frac{\mathbf{x}-\mathbf{\mu}_i}{||\mathbf{x}-\mathbf{\mu}_i||}} $$

A further L\textsubscript{2} normalization step is performed at the vector level, i.e. $ \mathbf{\hat{v}} = \frac{\mathbf{\dot{v}}}{||\mathbf{\dot{v}}||}$.

\begin{figure*}
\includegraphics[scale=0.56]{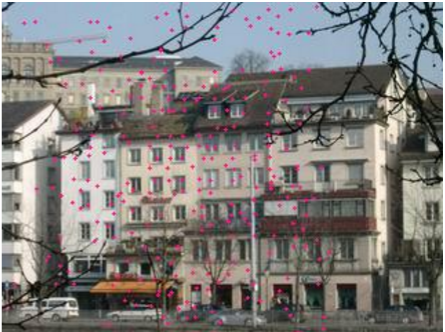}
\includegraphics[scale=0.56]{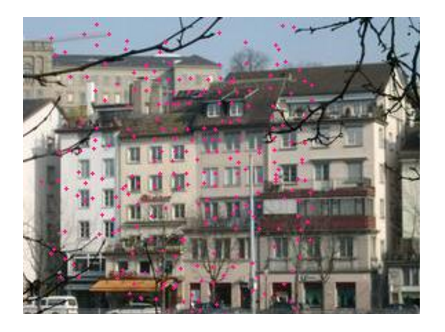}
\caption{An example of image from the ZuBuD dataset. The left image is the original query where 424 features are detected, while the right image shows the cropped verison with 367 featured retained.}
\label{fig:locVLAD}
\end{figure*}

One of the weakness of classical VLAD in landmark recognition is due to the noisy features corresponding to distractors such as trees, traffic signs, cars, people, etc.
The proposed variant of VLAD, called locVLAD (location-aware VLAD), tackles this problem by reducing the influence of features found at the borders of the image. One important point to make is that we do not simply remove features at the image borders, but fuse them at the VLAD descriptor level.
The VLAD procedure described above is performed twice (only on test phase), one considering the whole image, and the other considering the images cropped of a certain percentage.
It is quite straightforward (but important) to notice that, by repeating the whole pipeline, the feature set $x$ and, therefore, the VLAD vectors will be different. For instance, Fig. \ref{fig:locVLAD} shows how the detected features change.
Once the two VLAD vectors (denoted with $\mathbf{\dot{v}}$ and $\mathbf{\dot{v}_{cropped}}$) are computed, the locVLAD vector is simply obtained by averaging them:

\begin{equation} \label{locVLADequation}
\mathbf{\dot{v}_{locVLAD}} = \frac{\mathbf{\dot{v}}+\mathbf{\dot{v}_{cropped}}}{2}
\end{equation}

There are two main parameters to account for: the weights for the two vectors $\mathbf{\dot{v}}$ and $\mathbf{\dot{v}_{cropped}}$, and the percentage of borders to be cropped. Regarding the former, we performed different tests and realized that the best results are obtained by an equal weight for the two vectors, as shown in equation \ref{locVLADequation}. The second parameter depends on the resolution of the images in the dataset and will be discussed in the experiments.
The rationale behind our proposal is that the most important features (useful to recognize the landmark) are located near the center of the image, whereas the distractors are often at the border of the image (see, for instance, Fig. \ref{fig:locVLAD}).

%We tested also this method with different weights and we obtain that the best values are $0.5$ for both the descriptors.

%The dimension of the cropped image is a parameter, that depends on the used dataset. For ZuBuD and ZuBuD+ the cropped images are the size of 90\% compared to the original, instead for Holidays the size is reduced from 100\% to 70\%. This difference of cropping images is due to the resolution of the two image set: the Holidays images have a resolution higher than ZuBuD.  

%The improvement on accuracy recognition of this technique is due to that the important feature for the recognition, usually are in the center of the images and in the border usually there are noisy features.

The above rationale  might not be always true, i.e. some features at the image borders might be useful. For this reasone we average both the cropped and not-cropped VLAD descriptors. However, locVLAD procedure is not applied to the database images. Although this could be reasonable, experiments demonstrate that applying it also on the database images decreases the recognition accuracy. This behaviour can be explained by the fact that the database contains different views of the same landmark, also zoomed views. In these latter cases the significant features are located at the borders too and should not be removed. Therefore, the best results are achieved by applying locVLAD encoding on the query images only.

%Obviously, not in all images the features in the border of the image are noise, hence to generalize, because it does not say if the feature in the border are noise, this method it apply the mean of the VLAD descriptors: the original and the cropped image.
%Although this method improves the results, the application of locVLAD also on database images decreases the results because usually these pictures are different view of the same landmark and always there are also zoomed image, that do not contain any noisy features and the removed features are useful features for the recognition process.

%As a consequence the correct application of this method is only during the VLAD encoding phase of the query images.

%We tested also locVLAD with different VLAD descriptors: VLAD of the scaled image, VLAD of the rotated image, but no one improves the results.

\section{Experimental results}

In order to evaluate the accuracy of the proposed embedding technique with respect to the state of the art, we run experiments on public datasets and employing standard evaluation metrics. 

\subsection{Datasets and metrics}

The performance is measured on two public image datasets: ZuBuD and Holidays.

\textbf{ZuBuD} \cite{zubud} is composed by 1005 images of the size of 640x480 (or 480x640 if they are rotated), subdivided in 201 classes, about the building of Zurich. The query images are 115 and they are of the size of 320x240. Not all the classes are represented in the query set, but only 85.

\begin{figure*}
\includegraphics[scale=0.3]{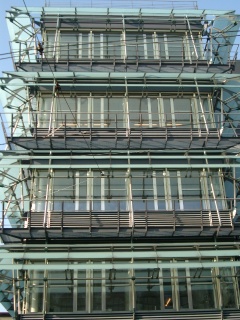}
\includegraphics[scale=0.3]{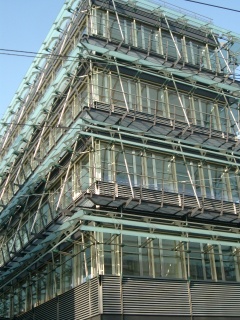}
\includegraphics[scale=0.3]{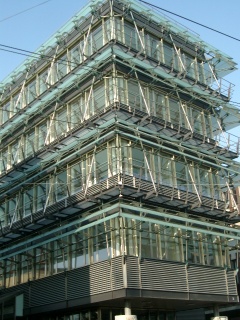}
\includegraphics[scale=0.3]{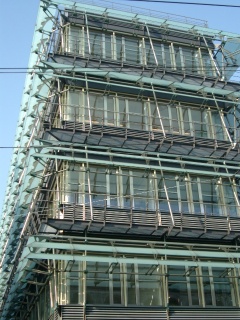}
\includegraphics[scale=0.3]{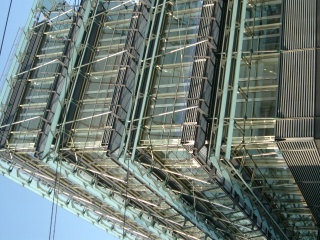}
\caption{Example of data added in the query set of ZuBuD.}
\label{ZuBuD+}
\end{figure*}

Since not all the classes are represented in the query set, in order to have a more balanced dataset, we created a new version of ZuBuD, called \textbf{ZuBuD+} (available at \url{http://implab.ce.unipr.it/?page\_id=194}). While keeping the database images unchanged, we extended the query images by randomly selecting images of the missing classes and transforming them by resizing and rotating them with ±90°. The total number of query images is raised to 1005 and all the classes are equally represented (5 queries per class). Fig. \ref{ZuBuD+} shows some examples of the newly-added query images.
According to \cite{zhang2015probabilistic}, to increase the recognition accuracy, we resized the database images to 320x240 (or 240x320 if they are rotated) during the creation of VLAD vectors for the database images.

%\textbf{ZuBuD+} is an extended version of ZuBuD, now available to all \footnote{http://implab.ce.unipr.it/?page\_id=194}. The database images are the same of ZuBuD and the query images are now 1005. All the classes are represented in equal manner. The added images in the query set are random selected from the database images of the missing relative class and transformed through: resize or rotation(±90°) and resize. In figure \ref{ZuBuD+} there are showed some pictures of images added to the new dataset.

\textbf{Holidays} \citep{jegou2008hamming} is composed of 1491 high-resolution images representing the holiday photos of different locations and objects, subdivided in 500 classes. The database images are 991 and the query images are 500, one for every class.

\textbf{Vocabulary creation.} The vocabulary on both ZuBuD and ZuBuD+ (because they have the same training images) is created by using all the features detected (about 208k features) since the images are few and of limited resolution. Conversely, Holidays is a larger dataset with higher resolution images and using all the features would result in a very large vocabulary. Therefore, we downsampled the number of features by randomly selecting 1/5 of the detected features (1.84 M out of the total features). On the downsampled set of features, K-Means++ (an approximated version of K-Means clustering for NP-hard problems) \cite{arthur2007k} is applied. The use of less features has a twofold motivation: first, it reduces computational time; second, it allows to reduce the chance of overfitting problem by supposedly avoiding to include features of the query images in the vocabulary. Finally, by randomly selecting the features to be retained (instead of selecting the first N features), we can avoid to overtrain on a particular patch.

%We choosed randomly $1/5$ of feature detected (1.84M of features than 9M of total feature detected ) and then apply K-Means++, that is an approximated version of K-Means clustering for NP-hard problems. This operation reduces the computational time during the phase of vocabulary creation and it does not bring in overfitting because the vocabulary is not generated on the feature of the query images. The random choice of the features, instead of the choice of the first $N$ features detected, reduces the possibility of overtraining on a particular patch.

\textbf{Size of cropped images.} As mentioned in the previous chapter the locVLAD approach is based on a mean of two VLAD descriptors: the VLAD on the original image and the one on the cropped image. The size of the cropped image have different values for every dataset used in the experiments. Before getting the best results, we tried different values. The best values are shown in Table \ref{tab:croppedValues}.

\begin{table}
\footnotesize
  \caption{Size of cropped images used for the application of locVLAD}
  \label{tab:croppedValues}
  \begin{tabular}{ccc}
    \toprule
    Dataset & Size of cropped image  \\
    \midrule
    ZuBuD and ZuBuD+ & 90\% of the original images' size  \\
    Holidays & 70\% of the original images' size  \\
    \bottomrule
  \end{tabular}
\end{table}

\textbf{Metrics.} Standard evaluation metrics include the mean average precision (mAP) or some ranking-based metrics, such as Top1 or 5xRecall@Top5 (average of how many times the correct image is in the top 5 results in the ranking). For the Holidays dataset we used the mAP provided by the corresponding evaluation tool, whereas for ZuBuD we prefered the ranking-based metrics which have been used in the compared works.
%On Holidays the image retrieval is measured by mean average precision (mAP) because there is provided an evaluation tool, that allow to calculate this metric. But on ZuBuD there is any tool; hence we calculate the accuracy of top1 retrieval and the 5 x recall@top5, that is the mean of how many images of the 5 retrievals are in the top 5 of the relative query. 

\textbf{Distance.} In order to compare a query image with the database, a $L_{2}$ distance is employed. An alternative distance is the cosine similarity, but results are similar and the computation is slower than the $L_{2}$ distance.
%During the retrieval to compare the query with the database is executed a brute force computation, using $L_{2}$ distance. We tested also the cosine similarity and we obtain the same results. Then we use $L_{2}$ distance because the calculation of the formula is faster than in the case of the cosine similarity.

\textbf{Implementation.} In term of actual implementation, the detector and descriptor used is SiftGPU \cite{siftgpu07wu}, that runs on GPU on a NVIDIA GeForce GTX 1070 mounted on a computer with a 8-core 3.40GHz CPU.

\subsection{Results on ZuBuD+}

\begin{table}
\footnotesize
  \caption{Comparison with the state of the art on ZuBuD and ZuBuD+ dataset}
  \label{tab:resultZuBuD}
  \begin{tabular}{ccccc}
    \toprule
    Method & Descriptor size & Top1 & 5xRecall@Top5 \\
    \midrule
    Tree histogram (ZuBuD) \cite{chen2009tree} & 10M & 98.00\% & - \\
    Decision tree (ZuBuD) \cite{fritz2006mobile} & n/a & 91.00\% & - \\
    Sparse coding (ZuBuD) \cite{zhang2015probabilistic} & 8k*64+1k*36 &  - & 4.538\\
    VLAD (ZuBuD) \cite{jegou2010aggregating} & 4281*128 & 99.00\% & 4.416 \\ 
    VLAD (ZuBuD+) \cite{jegou2010aggregating} & 4281*128 & 99.00\% & 4.526 \\ 
    locVLAD (ZuBuD) & 4281*128 & \textbf{100.00\%} & 4.469\\
    locVLAD (ZuBuD+) & 4281*128 & \textbf{100.00\%} & \textbf{4.543}\\
    \bottomrule
  \end{tabular}
\end{table}

We run the first set of experiments on the balanced dataset ZuBuD+. We compared the proposed locVLAD with standard VLAD as baseline. Moreover, we also compared our approach with published results on ZuBuD from three state-of-the-art papers. The tree histogram approach proposed in \cite{chen2009tree} uses a vocabulary of $10M$ visual words. Instead, the approach presented in \cite{fritz2006mobile} is based on decision trees and the i-SIFT detector described above. Finally, the third method compared \cite{zhang2015probabilistic} is based on sparse coding. 
In terms of vocabulary size (which an important parameter to compare having a direct effect on the recognition accuracy and the computational complexity of the approach), the first method \cite{chen2009tree} uses very large vocabulary so they can not be fairly compared with our (which is much more compact) and the second \cite{fritz2006mobile} does not use a vocabulary. However, as shown in Table  \ref{tab:resultZuBuD}, the locVLAD approach can obtain superior results on top1 even using a smaller vocabulary ($4281$ wrt $10M$ for \cite{chen2009tree}). Conversely, in the case of sparse coding in \cite{zhang2015probabilistic}, a vocabulary composed of $8k$ $64-D$ SURF descriptors plus $1k$ color descriptors is used. In order to perform a fair comparison, also locVLAD is tested with a vocabulary of the same size, i.e $64*8k+36*1k = 548k$. Since we are using $128-D$ SIFT descriptors, this means about $4281$ visual words.

\begin{figure}
\includegraphics[scale=0.5]{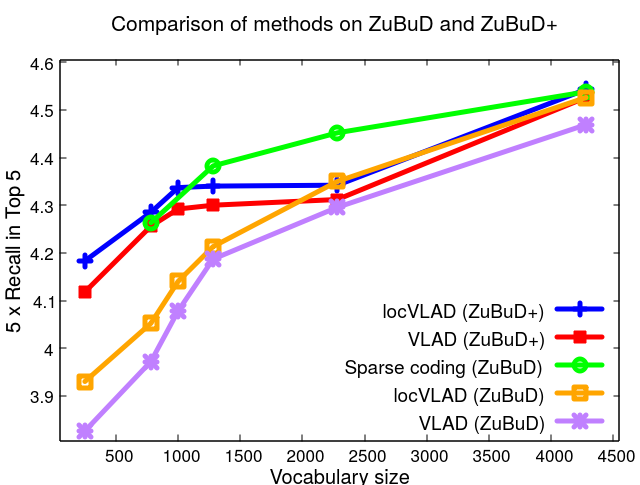}
\caption{We compare VLAD and locVLAD with the method based on sparse coding, using different vocabulary size on ZuBuD and ZuBuD+.}
\label{fig:graphZuBuD}
\end{figure}

Table \ref{tab:resultZuBuD} shows the results of locVLAD compared with baseline VLAD and sparse coding \cite{zhang2015probabilistic} with the same vocabulary size and on 5xRecall@Top5, while the comparison with \cite{chen2009tree, fritz2006mobile} is done at completely different sizes and on Top1. For locVLAD both results on ZuBuD and ZuBuD+ are shown. It is evident that our method outperforms \cite{chen2009tree, fritz2006mobile} and baseline VLAD, and that for the first two it is also using a much small vocabulary. However, our results compared with sparse coding are slightly worse when applied on ZuBuD. This can be explained by the unbalance in the query set described above. When applied on ZuBuD+, locVLAD outperforms the sparse coding results. This is also confirmed in Fig. \ref{fig:graphZuBuD} where we reported the comparison with the baseline VLAD and sparse coding at different vocabulary sizes.

\subsection{Results on Holidays}

In similar way, we run experiments on the Holidays dataset. As before, we compared with both the baseline VLAD and state-of-the-art methods. In this case, we compared again with \cite{zhang2015probabilistic} which uses sparse coding and max-pooling and with another paper \cite{cao2012weakly} using again sparse coding but with geometric pooling, i.e. local descriptors sharing good geometric consistency are pooled together to ensure a more precise spatial layouts.
As we did for ZuBuD, we set the vocabulary size to be comparable to that of the compared methods, i.e $4281$ for \cite{zhang2015probabilistic} and $20k$ for \cite{cao2012weakly}.

\begin{table}
  \caption{Comparison with the state of the art on Holidays dataset with different feature vectors size}
  \label{tab:resultHolidays1}
  \begin{tabular}{cccc}
    \toprule
    Method & Descriptor size & mAP\\
    \midrule
    Sparse coding \cite{zhang2015probabilistic} & 8k*64+1k*36 & 76.51\%\\
    VLAD \cite{jegou2010aggregating} & 4281*128 & 74.43\% \\
    locVLAD & 4281*128 & \textbf{77.20\%}\\
    \hline
    Sparse coding \cite{cao2012weakly} & 20k*128  & 79.00\% \\
    VLAD \cite{jegou2010aggregating} & 20k*128 & 78.78\%\\
    locVLAD & 20k*128 & \textbf{80.89\%}\\
    \bottomrule
  \end{tabular}
\end{table}

In fact, in this last case, the number $K$ of visual words is set to $20k$. Table \ref{tab:resultHolidays1} shows the results and clearly demonstrates that the proposed locVLAD achieves better mAP than the other methods.

\begin{figure}
\includegraphics[scale=0.5]{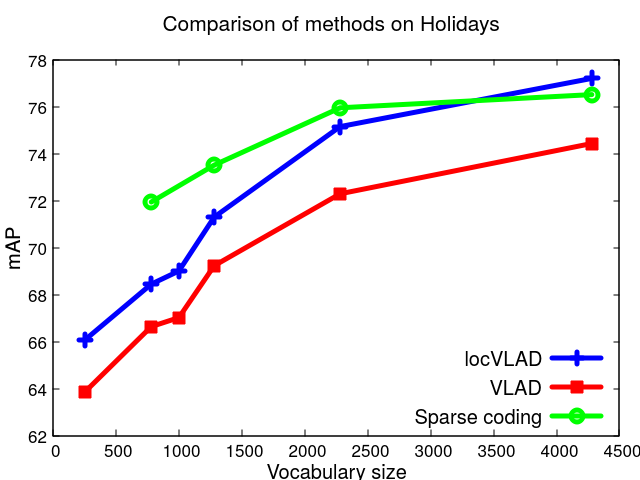}
\caption{We compare VLAD and locVLAD with the method based on sparse coding, using different vocabulary size on Holidays.}
\label{fig:graphHolidays}
\end{figure}

\begin{figure}
\includegraphics[scale=0.5]{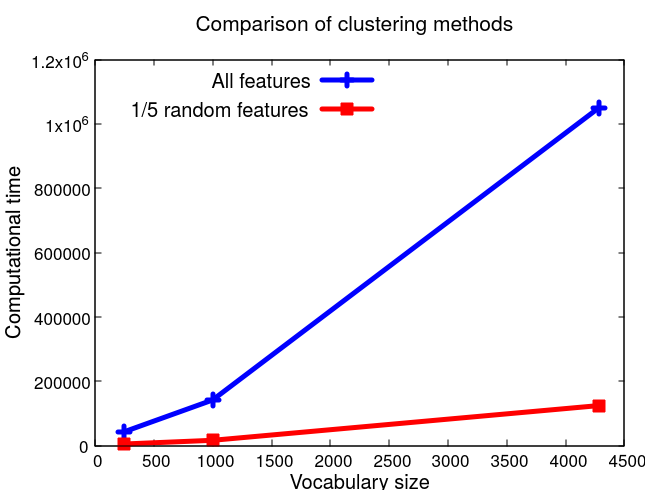}
\caption{Computational time in the clustering phase on Holidays dataset. The vocabulary size choosen are: $250$, $1000$ and $4281$.}
\label{fig:vocabularyCreation}
\end{figure}

As we showed in Fig. \ref{fig:graphZuBuD}, Fig. \ref{fig:graphHolidays} shows the different values of mAP achieved by baseline VLAD, sparse coding \cite{zhang2015probabilistic} and locVLAD at different vocabulary sizes.
Finally, Fig. \ref{fig:vocabularyCreation} shows the computational time needed for clustering at different vocabulary sizes when all the features compared with the $1/5$ downsampled features are used. As expected, the computational time increases very quickly when using all the features.

\section{Conclusions}
This paper proposes a novel embedding technique for efficient and effective landmark recognition. The proposed locVLAD technique includes, at a certain degree, information on the location of the features, by mitigating the negative efects of distractors found at the image borders.
Experiments are performed on two public datasets, namely ZuBuD and Holidays, and demonstrates superior recognition accuracy wrt the state of the art. It is worth to note that on ZuBuD the method based on sparse coding in \cite{zhang2015probabilistic} slightly outperforms the proposed one. This is due to an unbalanced query set and, probably, on the use of color information (which is not used in our approach). However, the results on both a more balanced dataset (ZuBuD+) and on the other dataset (Holidays) show that our method works better than \cite{zhang2015probabilistic}, substantially confirming our above-reported explanation.

\textbf{Acknowledgments.} This work is partially funded by Regione Emilia Romagna under the \textquotedblleft Piano triennale alte competenze per la ricerca, il trasferimento tecnologico e l'imprenditorialità\textquotedblright .
%\end{document}  % This is where a 'short' article might terminate

\bibliographystyle{ACM-Reference-Format}
\bibliography{sigproc} 

\end{document}